\pdfoutput=1

\documentclass[11pt]{article}
\usepackage{acl}
\usepackage{times}
\usepackage{latexsym}
\usepackage[T1]{fontenc}
\usepackage[utf8]{inputenc}
\usepackage[T2A,LAE,T1]{fontenc}
\usepackage[arabic,USenglish]{babel}
\usepackage{microtype}
\usepackage{inconsolata}
\usepackage{graphicx}
\usepackage{listings}
\usepackage{tabularx}
\usepackage{booktabs}
\usepackage{todonotes}
\usepackage{multirow}
\usepackage{array}
\newcolumntype{P}[1]{>{\centering\arraybackslash}p{#1}}
\usepackage{caption}
\usepackage{tabularx}
\usepackage{subcaption}
\usepackage{url}

\usepackage{xcolor}
\definecolor{Purple}{RGB}{118, 42, 131}
\usepackage[listings]{tcolorbox}
\usepackage{rotating}
\usepackage{float}
\newcommand\nstrongcorrdatasets{8 }
\newcommand\nstrongvalidhypothesis{11 }
\newcommand\ndatasets{15 }
\newcommand\nlargepropmsadatasets{6 }
\newcommand\ntasks{6 }

\title{Estimating the Level of Dialectness Predicts Interannotator Agreement in Multi-dialect Arabic Datasets}

\author{
Amr Keleg, Walid Magdy, Sharon Goldwater\\
Institute for Language, Cognition and Computation \\
School of Informatics, University of Edinburgh \\
\texttt{a.keleg@sms.ed.ac.uk}, \texttt{\{wmagdy,sgwater\}@inf.ed.ac.uk}}

\begin{document}
\maketitle

\begin{abstract}

On annotating multi-dialect Arabic datasets, it is common to randomly assign the samples across a pool of native Arabic speakers. Recent analyses recommended routing dialectal samples to native speakers of their respective dialects to build higher-quality datasets. However, automatically identifying the dialect of samples is hard. Moreover, the pool of annotators who are native speakers of specific Arabic dialects might be scarce.
Arabic Level of Dialectness (ALDi) was recently introduced as a quantitative variable that measures how sentences diverge from Standard Arabic.
On randomly assigning samples to annotators, we hypothesize that samples of higher ALDi scores are harder to label especially if they are written in dialects that the annotators do not speak.
We test this by analyzing the relation between ALDi scores and the annotators' agreement, on \ndatasets public datasets having raw individual sample annotations for various sentence-classification tasks. We find strong evidence supporting our hypothesis for \nstrongvalidhypothesis of them.
Consequently, we recommend prioritizing routing samples of high ALDi scores to native speakers of each sample's dialect, for which the dialect could be automatically identified at higher accuracies. 
\end{abstract}

\section{Introduction}

Arabic is spoken natively by over 420 million people and is an official language of 24 countries \cite{bergman-diab-2022-towards}, making it an important language for NLP systems. However, NLP for Arabic faces a major challenge in that user-generated text is typically a mixture of Modern Standard Arabic (MSA)---the standardized variant that is taught in schools and used in official communications and newspapers---and regional variants of Dialectal Arabic (DA), which are used in everyday communications, including both speech and social media text \cite{79835034091e4ec8a327712e27b1b785}.\footnote{Refer to §\ref{sec:msa_da_code_mixing} of the Appendix for a further discussion about the relationship between MSA and DA.} While MSA can be largely understood by most Arabic speakers, the different variants of DA are not always fully mutually intelligible.

Despite this mutual unintelligibility, a common practice when developing datasets for multi-dialect Arabic NLP is to
randomly recruit annotators without regard to their dialect.
However, routing dialectal content to speakers of a different dialect for annotation or moderation can present real problems. For example, it has been shown to contribute to unjust online content moderation of DA \cite{BSR_report}, and racially biased toxicity annotation in American English varieties \cite{sap-etal-2022-annotators}. Two recent studies of multi-dialect DA annotation showed that for annotating hate speech or sarcasm, respectively, annotators were more lenient (for hate speech) and more accurate (for sarcasm) when annotating sentences in their native dialect \cite{bergman-diab-2022-towards, abu-farha-magdy-2022-effect}. 
The authors of both studies made the same recommendation for creating new Arabic datasets, namely to first identify the dialect of each sample and then route it to appropriate annotators.

This recommendation is theoretically appealing, but presents practical difficulties since automatic dialect identification (DI) is challenging \cite{abdulmageed2023nadi}, and existing systems assume a single correct label when in fact some texts can be natural in different dialects \cite{Keleg2023ArabicDI,olsen-etal-2023-arabic}.
Moreover, the representation of native speakers of the different Arabic dialects on crowdsourcing sites might be skewed \cite{mubarak2016demographic}. Therefore, recruiting native speakers of some Arabic dialects might be challenging, given the tough conditions of the countries in which these dialects are spoken.

In this paper, we address these challenges by building on recent work by \citet{keleg2023aldi}, who presented a system for estimating Arabic Level of Dialectness (ALDi)---i.e., the degree to which a sentence diverges from MSA, on a scale from 0 to 1. We hypothesize that as sentences with low ALDi scores do not diverge much from MSA, they can still be understood and accurately annotated by most Arabic speakers, while this will be less true for sentences with high ALDi scores.
If our hypothesis holds, then annotation can be made more efficient while maintaining accuracy, by routing samples with low ALDi scores to speakers of any dialect. Only high-ALDi samples need to be routed to native speakers of the appropriate dialect.

We test our hypothesis by investigating the impact of ALDi score on interannotator agreement for \ndatasets publicly released datasets annotated for \ntasks different sentence-classification tasks.\footnote{Instructions to replicate the experiments can be accessed through \url{https://github.com/AMR-KELEG/ALDi-and-IAA}} We confirm that for most tasks and datasets, higher ALDi scores correlate with lower annotator agreement. A notable exception is the dialect identification (DI) task, where higher ALDi scores correlate with \emph{higher} agreement, presumably because it is easier to identify a single dialect for sentences that are strongly dialectal. This finding is encouraging for annotation routing, since automatic DI systems may also have higher accuracy on these sentences. We conclude that a combination of automatic ALDi scoring, followed by DI and annotator routing only for high-ALDi sentences, is a promising strategy for annotating multi-dialect Arabic datasets.

\section{Methodology}
\label{sec:methodology}
\begin{table*}[t]
    \centering
    \scriptsize
    \begin{tabular}{p{3cm}p{1.5cm}p{1cm}p{8cm}}
 \textbf{Dataset}  &         \textbf{Task (\# labels)}  &  \textbf{\%ALDi<0.1}  &  \textbf{Description} \\
        \midrule
 Deleted Comments Dataset (DCD) \cite{mubarak-etal-2017-abusive}  &         Offensive (3)   &  62.57\%  & About 32K deleted comments from \url{aljazeera.com}. Confidence scores for the majority vote of 3 annotations are provided. \\
        \midrule
 MPOLD \cite{chowdhury-etal-2020-multi}  &         Offensive (2)  &  27.82\%  &   4000 sentences interacting with news sources, sampled from Twitter, Facebook, and YouTube, annotated three times.\\
        \midrule
 YouTube Cyberbullying  &         Offensive (2)   &  10.24\%  &  \multirow{2}{8cm}{15,050 comments and replies to 9 YouTube videos labeled by 3 annotators (Iraqi, Egyptian, Libyan).}\\
 (YTCB) \cite{ALAKROT2018174}  &          &  \\
        \midrule
ASAD \cite{alharbi2021asad}  &         Sentiment (3)   &  35.63\%  &  95,000 tweets with a skewed representation toward the Gulf area and Egypt.\\
        \midrule
ArSAS \cite{d3cb00d902eb44a0a0d01b794e862787}  &         Sentiment (4)  &  57.45\%  &  \multirow{2}{8cm}{21,064 tweets related to a pre-specified set of entities or events, with confidence scores for the majority votes across three annotations per sample.}\\
  &         Speech Act (6)  &  \\
    \midrule
  ArSarcasm-v1  &         Dialect (5)  &  57.44\%  &  \multirow{2}{8cm}{10,547 tweets, sampled from two different Sentiment Analysis datasets: ATSD \cite{nabil-etal-2015-astd}, SemEval2017 \cite{rosenthal-etal-2017-semeval}, reannotated for Sentiment, Dialect, and Sarcasm.}\\
 \cite{abu-farha-magdy-2020-arabic}  &         Sarcasm (2)  &  \\
  &         Sentiment (4)  &  \\
        \midrule
 Mawqif  &         Sarcasm (2)  &  58.04\%  &  \multirow{3}{8cm}{4,121 tweets about "COVID-19 vaccine", "digital transformation", or "women empowerment" annotated separately for stance and sentiment/sarcasm till the label confidence reaches 0.7 (min. 3 annotators) or 7 annotators label the sample.}\\
 \cite{alturayeif-etal-2022-mawqif}  &         Sentiment (3)   & 58.04\% &   \\
  &         Stance (3)  &   57.99\%  &  \\
        \midrule
 iSarcasm's test set  &         Dialect (5)  &  30.5\%  &  \multirow{2}{8cm}{200 sarcastic sentences provided by crowdsourced authors and 1200 non-sarcastic tweets from ArSarcasm-v2 \cite{abu-farha-etal-2021-overview} reannotated 5 times.}\\
 \cite{abu-farha-etal-2022-semeval}  &         Sarcasm (2)  &  \\
        \midrule
 DART \cite{alsarsour-etal-2018-dart}  &         Dialect (5)  & 0.8\%  &  24,279 tweets with distinctive dialectal terms annotated three times for the dialectal region. Samples of complete disagreement are not in the released dataset.\\
        \bottomrule
    \end{tabular}
    \caption{
   The datasets included in our study. All datasets have three annotations per sample, except for iSarcasm (5 annotations/sample) and Mawqif (3 or more annotations/sample).
   For the labels used in each dataset and the proportion of each label, see Table~\ref{tab:datasets_detailed}.
   For some datasets, there is a discrepancy between the number of samples listed in the paper and the raw data files with individual labels (See \S\ref{sec:discrepancy} of the Appendix).
}
    \label{tab:datasets}
\end{table*}

\paragraph{Data} We study the impact of ALDi scores on the annotators' agreement for publicly released Arabic datasets.
We analyze datasets satisfying the following criteria:
\vspace{-0.3cm}
\begin{itemize}
    \setlength{\parskip}{0pt}
    \setlength{\itemsep}{0pt}
    \item \textbf{Language}: Mixture of MSA and DA.
    \item \textbf{Variation}: Targeting multiple variants of DA.
    \item \textbf{Annotators}: Speakers of different variants of DA that are randomly assigned to the samples.
    \item \textbf{Tasks Setup}: Sentence-level classification.
    \item \textbf{Released Labels}: Individual annotator labels or the percentage of annotators agreeing on the majority-vote label.\footnote{For some datasets, the percentage of annotators agreeing on the majority vote is weighted by their performance on the annotation quality-assurance test samples. This distinction is irrelevant to our study, where we only consider whether all annotators agreed or not.}
\end{itemize}

We searched for datasets on Masader, a community-curated catalog of Arabic datasets \cite{alyafeai2021masader, altaher2022masader}. Each dataset on Masader has a metadata field for the variants of Arabic included. We discarded the datasets that only included MSA samples, and manually inspected the remaining 151. After identifying 28 potential datasets that satisfy the criteria above, we contacted the authors of the datasets that do not have the individual annotations publicly released. Eventually, we had \ndatasets datasets to analyze, listed in Table \ref{tab:datasets}, covering: Offensive Text Classification, Hate Speech Detection, Sarcasm Detection, Sentiment Analysis, Speech Act Detection, Stance Detection, and Dialect Identification.

\paragraph{Analysis}
For each dataset, we compute the Arabic Level of Dialectness (ALDi) score for each annotated sample (sentence) using the Sentence-ALDi model \cite{keleg2023aldi}, which returns a score from 0 (MSA/non-dialectal) to 1 (strongly dialectal). To investigate the effect of ALDi on annotator agreement, we bin the samples by their ALDi score into 10 bins of width 0.1. We compute \textit{\% full agree}, the percentage of samples in that bin for which all the annotators agreed on a single label. We employ Pearson's correlation coefficient to analyze the relation between ALDi (represented by each bin's midpoint ALDi score) and \textit{\% full agree}, and also report the slope of the best-fitting line as a measure of the effect size.\footnote{The exact values of the slopes and correlation coefficients depend on the number of bins. However, we got similar qualitative results on using 4 or 20 equal-width bins. 10 bins are enough to check if trends are non-linear while keeping a reasonable number of samples in the smallest bins. We also fitted logistic regression (\textit{logreg}) models using ALDi as a continuous variable and a binary outcome \textit{Full Agreement (Yes/No)} for each sample. Both analysis tools reveal similar patterns (See Appendix \S\ref{sec:other_analysis}) but the binning method provides useful additional visualization.}
As aforementioned, our initial hypothesis is that \textit{\% full agree} negatively correlates with high ALDi scores.
 
\section{Results and Discussion}
\label{sec:results}

\begin{figure*}[t!]
    \centering
    \scriptsize{\textbf{Sentiment Analysis}}\\
    \begin{flushleft}
        \vspace{-0.65cm}
        \rule{16.5cm}{0.1mm}
        \vspace{-0.48cm}
    \end{flushleft}
    
    \begin{subfigure}[t]{0.25\textwidth}
        \centering
        \includegraphics[trim=0 0.25cm 0 0.2cm, clip]{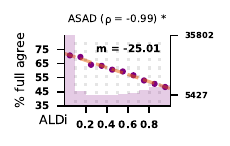}
    \end{subfigure}%
    ~
    \begin{subfigure}[t]{0.25\textwidth}
        \centering
        \includegraphics[trim=0 0.25cm 0 0.2cm, clip]{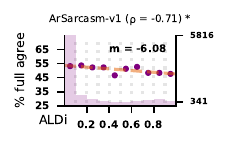}
    \end{subfigure}%
    ~
    \begin{subfigure}[t]{0.25\textwidth}
        \centering
        \includegraphics[trim=0 0.25cm 0 0.2cm, clip]{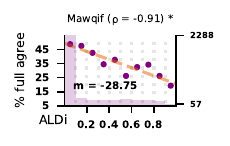}
    \end{subfigure}%
    ~
    \begin{subfigure}[t]{0.25\textwidth}
        \centering
        \includegraphics[trim=0 0.25cm 0 0.2cm, clip]{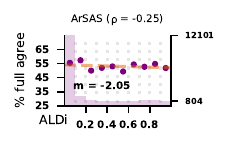}
    \end{subfigure}%

    \vfill
    
    \scriptsize{\textbf{Sarcasm Detection}}\\
    \begin{flushleft}
        \vspace{-0.4cm}
        \rule{12cm}{0.1mm}
    \end{flushleft}

    \begin{subfigure}[t]{0.25\textwidth}
        \centering
        \includegraphics[trim=0 0.25cm 0 0.2cm, clip]{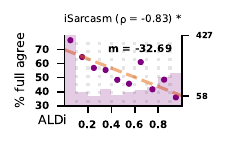}
    \end{subfigure}%
    ~
    \begin{subfigure}[t]{0.25\textwidth}
        \centering
        \includegraphics[trim=0 0.25cm 0 0.2cm, clip]{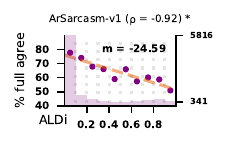}
    \end{subfigure}%
    ~
    \begin{subfigure}[t]{0.25\textwidth}
        \centering
        \includegraphics[trim=0 0.25cm 0 0.2cm, clip]{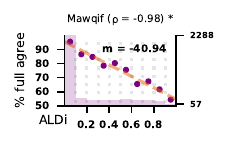}
    \end{subfigure}%
    \begin{subfigure}[t]{0.25\textwidth}
        \centering
        \vspace{-2.4cm}
        \scriptsize{\underline{\textbf{Speech Act Detection}}}
        \includegraphics[trim=0 0.25cm 0 0.2cm, clip]{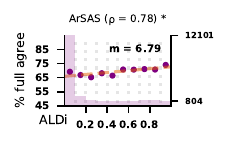}
    \end{subfigure}%
    \vfill

    \scriptsize{\textbf{Offensive Text Classification and Hate Speech Detection}}
    \begin{flushleft}
        \vspace{-0.3cm}
        \rule{12cm}{0.1mm}
    \end{flushleft}

    \begin{subfigure}[t]{0.25\textwidth}
        \centering
        \includegraphics[trim=0 0.25cm 0 0.2cm, clip]{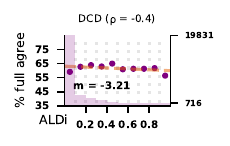}
    \end{subfigure}%
    ~
    \begin{subfigure}[t]{0.25\textwidth}
        \centering
        \includegraphics[trim=0 0.25cm 0 0.2cm, clip]{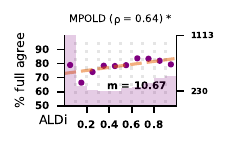}
    \end{subfigure}%
    ~
    \begin{subfigure}[t]{0.25\textwidth}
        \centering
        \includegraphics[trim=0 0.25cm 0 0.2cm, clip]{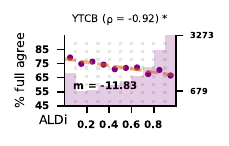}
    \end{subfigure}%
    ~
    \begin{subfigure}[t]{0.25\textwidth}
        \centering
        \vspace{-2.4cm}
        \scriptsize{\underline{\textbf{Stance Detection}}}
        \includegraphics[trim=0 0.25cm 0 0.2cm, clip]{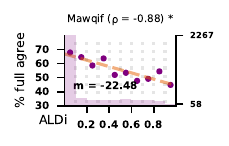}
    \end{subfigure}%
    \vfill
    
    \scriptsize{\textbf{Arabic Dialect Identification}}
    \begin{flushleft}
        \vspace{-0.3cm}
        \rule{12cm}{0.1mm}
    \end{flushleft}

    \begin{subfigure}[t]{0.25\textwidth}
        \centering
        \includegraphics[trim=0 0.25cm 0 0.2cm, clip]{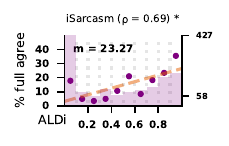}
    \end{subfigure}%
    ~
    \begin{subfigure}[t]{0.25\textwidth}
        \centering
        \includegraphics[trim=0 0.25cm 0 0.2cm, clip]{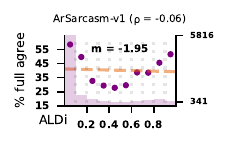}
    \end{subfigure}%
    ~
    \begin{subfigure}[t]{0.25\textwidth}
        \centering
        \includegraphics[trim=0 0.25cm 0 0.2cm, clip]{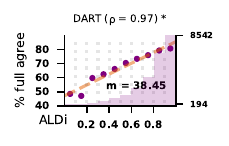}
    \end{subfigure}%
    ~
    \begin{subfigure}[t]{0.25\textwidth}
        \centering
        \phantom{\includegraphics[trim=0 0.25cm 0 0.2cm, clip]{new_plots/DART_10_dialect_merged.pdf}}
    \end{subfigure}

    \caption{Scatter plots showing the relationship between binned ALDi scores (x-axis) and the percentage of samples with full annotator agreement (y-axis). The histogram represents the \# of samples per bin (with min and max values for any bin labeled on the right-hand axis). The slope of the best-fitting line ($m$) is shown, and to enable visual comparison of slopes, all plots have the same y-axis scale (possibly shifted up or down).\\ \textbf{Note:} Statistically significant (p<0.05) correlation coefficients ($\rho$) are marked with *.}
    \label{fig:agreement_percentages_new}
\end{figure*}

\begin{table*}[t]
    \centering
    \small
        \begin{tabular}{P{2cm}P{6cm}P{3cm}cP{2cm}}
            \textbf{Dataset} & \textbf{Sample} & \textbf{Translation of the Underlined Cue} & \textbf{ALDi} & \textbf{Individual Labels} \\
            \midrule
            \textbf{YTCB} Hatespeech Detection & \scriptsize{\underline{\AR{يا حشرة}}  \AR{كاع واش كاينة و لا حتا حاس بيك} \AR{المغرب ما عرفكش}} & \underline{you insect} & 0.98 & HateSpeech(3x) \\
            \cmidrule{2-5}
            & \scriptsize{\AR{عمر تروح فدوة للتراب الي اتدوس علية شيعية العراق يا عرب \underline{\AR{يا جرب}}}} & \underline{you scabies} & 0.81 & HateSpeech(3x)\\
            \cmidrule{2-5}
            & \scriptsize{\AR{ماكان ناقص كاظم الا بلبل عفآ \underline{\AR{جحش الخليج}} يحكي عنو هههههههههههههههه	}} & \underline{Gulf's colt} & 0.98 & HateSpeech(3x)\\
            \midrule
            \textbf{ArSarcasm-v1} Sentiment Analysis & \scriptsize{\AR{ريكي مارتن كان قبل يعجبني بس بعد كلامه \underline{\AR{حسيته مقرف}} يلوع بالجبد}}& \underline{I found him disgusting}& 0.81 & Negative(2x) Positive(1x)\\
            \cmidrule{2-5}
            & \scriptsize{\AR{\#اطفال دول ولا مش اطفال ؟؟؟ظظ يابتوع \#المدارس \underline{\AR{ههههههههههههههههههههه	}}}} & \underline{hahaha} & 0.94& Positive(2x) Negative(1x)\\
            \bottomrule
        \end{tabular}
    \caption{Five qualitative samples of high ALDi scores. The underlined segments represent the cues that the annotators might have used to choose a label even if they do not fully understand the sentence. Despite the presence of these cues, the annotators still disagreed on labeling the last two samples. We only provide translations for the underlined cue segments.}
    \label{tab:qualitiative_samples}
\end{table*}

We use scatter plots to visualize the relation between \textit{\% full agree} and ALDi on the studied datasets, as shown in Figure~\ref{fig:agreement_percentages_new}. Additionally, the histograms of samples across the different bins indicate the dialectal content within the dataset. As per Table~\ref{tab:datasets}, \nlargepropmsadatasets datasets out of the \ndatasets have more than 50\% of the samples with ALDi scores less than 0.1, which are expected to be written in MSA. However, we found that the overall trends depicted in Figure~\ref{fig:agreement_percentages_new} will not be affected if we discard these samples with low ALDi scores and only focus on the rest.

\paragraph{For non-DI tasks, ALDi negatively correlates with agreement.}
Inspecting the trends depicted in Figure~\ref{fig:agreement_percentages_new}, strong negative Pearson's correlation coefficients exist for \nstrongcorrdatasets out of the 12 datasets for the non-DI tasks (sentiment analysis, sarcasm, hate speech, and stance detection).
Both the trends (quantified by the slope \textit{m}) and the correlation coefficients for most of the tasks indicate that the percentage of samples for which all the annotators assign the same label decreases as the ALDi scores increase, often by a large margin.\footnote{Refer to Appendix §\ref{sec:arsas_trends} for a possible explanation of the unexpected trends of the ArSAS dataset.}
We notice different trends for DI that we elaborate on below.

\paragraph{For DI, agreement is lowest for mid-range ALDi scores (if MSA is a possible label) or low ALDi scores (if it is not).}

By definition, MSA sentences have an ALDi of 0, and normally the ALDi estimation model assigns them very low scores.

For the ArSarcasm-v1 and iSarcasm datasets, the set of labels for the DI task includes MSA (i.e., some sentences in these datasets are not dialectal). For both datasets, one notices high percentages of agreement scores for the bin having ALDi scores $\in [0, 0.1]$ (generally agreeing that the label is MSA). The percentages decrease for the few succeeding bins, before rising again for the bins with high ALDi scores. Sentences of high ALDi scores (e.g., $\in [0.8, 1]$) are expected to have multiple dialectal cues, which increases the chance of attributing them to a single dialect. For sentences of intermediate ALDi scores, annotators can agree a sentence is not in MSA, however, they would struggle to determine the dialect of the sentence, which is manifested as having lower percentages of the full agreement for these bins.

The authors of the DART dataset do not include MSA in the labels since they curated sentences with distinctive dialectal terms. This explains the low percentage of full agreement for the bin of ALDi scores $\in [0, 0.1]$, unlike the other two DI datasets. However, the pattern of having higher full agreement percentages for bins with higher ALDi scores still holds.

\section{Analysis of Trends by Class Label}
\label{sec:detailed_analysis}

A more nuanced analysis of the non-DI datasets can be done by splitting the samples according to their majority-vote labels (See Appendix \S\ref{sec:detailed_trends}).

The declining trend for agreement as ALDi scores increase is consistently salient for the negative class of the different tasks (i.e., \textit{Not Sarcastic}, \textit{Neutral} (sentiment), \textit{Non-offensive}). One explanation is that agreeing on one of these labels requires fully understanding the sample, which is expected to be harder for non-native speakers of the dialects of sentences with high ALDi scores. Moreover, the general usage of MSA for formal communications, and DA for more personal ones might be biasing the annotators. Further controlled analysis is required to investigate these intuitive explanations.

Conversely, the presence of specific words might be a strong cue for the label of the overall sentence, which might be useful for the positive classes (i.e., \textit{Sarcastic}, \textit{Positive/Negative}, \textit{Obscene/Offensive}). Consequently, annotators might be able to agree on a label for a sentence even if it contains some unintelligible segments.
Table~\ref{tab:qualitiative_samples} shows five examples, in which the annotators can use cues to label the whole sentence.
Consider the first offensive sentence in the Table. The presence of the MSA insult  \begin{footnotesize}\AR{يا حشرة}\end{footnotesize} (you insect) is enough to guess that the sentence is offensive, even if the remaining segment is not fully intelligible. Lastly, note that agreeing on a label does not imply it is accurate, especially when relying on cues for annotation.

\section{Conclusion and Recommendation}
Factors such as task subjectivity and vague guidelines could cause disagreement between annotators. For Arabic, we demonstrate that the Arabic level of dialectness of a sentence (ALDi), automatically estimated using the Sentence-ALDi model \cite{keleg2023aldi}, is an additional overlooked factor.

Analyzing \ndatasets datasets, we find strong evidence of a negative correlation between ALDi and the full annotator agreement scores for \nstrongcorrdatasets of the 12 non-Dialect Identification datasets. Moreover, for the 3  Dialect Identification datasets, we find that annotators have higher agreement scores for samples of higher ALDi scores, which by definition would have more dialectal features. The combination of more dialectal features in a sentence is more probable to be distinctive of a specific dialect. 

Previous research recommended routing samples to native speakers of the different Arabic dialects for higher annotation quality.
Our analysis indicates that a large proportion of 6 datasets are samples with ALDi scores < 0.1, which are expected to be MSA samples that can be routed to speakers of any Arabic dialect. Moreover, the lower agreement scores for samples with high ALDi scores show that extra care should be given to these samples. Dataset creators should prioritize routing high-ALDi samples to native speakers of the dialects of these samples, for which the dialects can be automatically identified at higher accuracy as these samples have more dialectal cues.

\section*{Limitations}
The trends we report validate our hypothesis. However, more thorough analyses need to be done to understand how ALDi affects each task given its unique nature. Knowing the demographic information about the annotators might have allowed for revealing deeper insights into how speakers of specific Arabic dialects understand samples from other dialects. However, this would have required running a controlled experiment re-annotating the \ndatasets datasets, which we hope future work will attempt.

Another potential extension to this work is to analyze the interannotator disagreement on annotating dialectal data for token-level tasks. To the best of our knowledge, all the publicly available token-level Arabic datasets are built by carefully selecting samples written in specific dialects and recruiting native speakers of each of these dialects to perform the annotation, after closely training them. However, even if a multi-dialect token-level dataset is annotated by randomly assigning the samples to speakers of different dialects, the analysis would require a new model to estimate the level of dialectness on the token level, since the \textit{Sentence-ALDi} model used here works at the sentence level.

Lastly, we acknowledge that there are multiple reasons for the annotators to disagree, which include the task's subjectivity, the annotators' background, and their worldviews \cite{uma2021learning}. However, these factors would have less impact on the annotators' disagreement if a sample is not fully intelligible.

\section*{Acknowledgments}

This work could not have been done without the help of the datasets' creators who have kindly agreed to share the individual labels for their datasets' samples. Thanks, Ibrahim Abu Farha, Nora Alturayeif, and Manal Alshehri. We also thank Hamdy Mubarak, Nuha Albadi, Nedjma Ousidhoum, and Hala Mulki for trying to help with finding the individual annotator labels for some of their datasets. Lastly, we appreciate the efforts of the anonymous ARR reviewers, action editors, and area chairs. Thanks for the insightful discussions and valuable suggestions.

This work was supported by the UKRI Centre for Doctoral Training in Natural Language Processing, funded by the UKRI (grant EP/S022481/1) and the University of Edinburgh, School of Informatics.

\bibliography{anthology,custom}

\appendix
\setcounter{table}{0}
\setcounter{figure}{0}
\renewcommand{\thetable}{\Alph{section}\arabic{table}}
\renewcommand{\thefigure}{\Alph{section}\arabic{figure}}

\section{Detailed Description of the Datasets}
\label{sec:discrepancy}

\begin{table*}[!h]
    \centering
    \scriptsize
    \begin{tabular}{p{3cm}p{1.5cm}p{1cm}P{4cm}P{4cm}}
 \textbf{Dataset}  &         \textbf{Task (\# labels)}  &  \textbf{Labels}  &  \textbf{Distribution of Majority-vote Labels} & \textbf{Dataset/Paper Discrepancy}\\
        \midrule
 Deleted Comments Dataset (DCD) \cite{mubarak-etal-2017-abusive}  & Offensive (3) & Confidence & Offensive (80.31\%) Clean (17.76\%) Obscene (1.58\%) No Majority (0.35\%) & - \\
        \midrule
 MPOLD \mbox{\cite{chowdhury-etal-2020-multi}}  &         Offensive (2)  & Individual & Non-Offensive (83.12\%) Offensive (16.88\%) & - \\
        \midrule
 YouTube Cyberbullying  &         Offensive (2)   & Individual & Not (61.38\%) HateSpeech (38.62\%) & - \\
 (YTCB) \cite{ALAKROT2018174}  &          &  \\
        \midrule
ASAD \cite{alharbi2021asad}  &         Sentiment (3)   & Individual & Neutral (67.83\%) Negative (15.33\%) Positive (15.19\%) No Majority (1.65\%) & \multirow{5}{4cm}{The authors shared with us the raw annotation file of which we analyze 100,484 samples with three annotations or more, as opposed to the 95,000 in the released dataset.}\\
& & & & \\
& & & & \\
& & & & \\
        \midrule
ArSAS \cite{d3cb00d902eb44a0a0d01b794e862787}  &         Sentiment (4)  & Confidence & Negative (35.38\%) Neutral (33.45\%) Positive (20.51\%) No Majority (6.07\%) Mixed (4.59\%) & \multirow{2}{*}{-}\\
  &         Speech Act (6)  &  Confidence & Expression (55.07\%) Assertion (38.63\%) Question (3.32\%) No Majority (1.81\%) Request (0.67\%) Recommendation (0.31\%) Miscellaneous (0.18\%) & \\
        \midrule
  ArSarcasm-v1 \mbox{\cite{abu-farha-magdy-2020-arabic}} &         Dialect (5)  & Individual &  msa (67.56\%) egypt (19.37\%) No Majority (5.83\%) gulf (3.61\%) levant (3.46\%) magreb (0.18\%) & \multirow{6}{4cm}{The samples in the raw annotation artifact shared by the authors has 10,641 samples, as opposed to the 10,547 samples in the released dataset.}\\
   &         Sarcasm (2)  & Individual & False (84.24\%) True (15.7\%) No Majority (0.06\%) & \\
  &         Sentiment (3)  & Individual &  neutral (49.45\%) negative (32.57\%) positive (14.58\%) No Majority (3.4\%)\\
        \midrule
 Mawqif \mbox{\cite{alturayeif-etal-2022-mawqif}} &         Sarcasm (2)  & Individual &  No (95.97\%) Yes (3.78\%) No Majority (0.25\%) & \multirow{6}{4cm}{{The authors annotated the same samples for sentiment/sarcasm and stance separately. This was done across 8 different annotation jobs (4 each), for which the authors shared the raw annotation files with us. The number of samples in these files is 4,093 for sentiment/sarcasm and 4,079 for stance, of which 3,942 and 3,909 have three or more annotations. The released dataset is reported to have 4,100 samples.}} \\
  & & & & \\
  & & & & \\
  &         Sentiment (3)   & Individual &   Positive (41.15\%) Negative (31.46\%) Neutral (22.68\%) No Majority (4.72\%) & \\
  & & & & \\
  & & & & \\
  & Stance (3) & Individual & Favor (60.5\%) Against (27.65\%) None (7.7\%) No Majority (4.14\%) & \\
  & & & & \\
        \midrule
 \mbox{iSarcasm's test set} \mbox{\cite{abu-farha-etal-2022-semeval}} &         Dialect (5)  & Individual & msa (32.29\%) nile (31.36\%) gulf (16.5\%) No Majority (15.79\%) levant (2.21\%) maghreb (1.86\%) & \multirow{4}{4cm}{The dataset having the individual annotator labels is released as an artifact accompanying the following paper \cite{abu-farha-magdy-2022-effect}.} \\
   &   Sarcasm (2)  & Individual &  0 (82.07\%) 1 (17.93\%) \\
        \midrule
 DART \cite{alsarsour-etal-2018-dart}  &         Dialect (5)  & Proportion  &  GLF (24.27\%) EGY (21.69\%) IRQ (21.64\%) LEV (16.22\%) MGH (16.18\%) & -\\
        \bottomrule
    \end{tabular}
    \caption{
   A detailed description of the distribution of the majority-vote labels and the data/paper discrepancies in the datasets with individual annotator labels included in our study.\\ \textbf{Note 1:} \textit{No Majority} means that multiple labels have the same majority number of votes for Individual/Proportion labels, and Confidence < 0.5 otherwise.\\ \textbf{Note 2:} Some of the samples of the \textit{ASAD}, \textit{ArSarcasm-v1}, \textit{Mawqif} datasets have more than 3 annotations, despite the fact the former two are supposed to have only three annotations per sample.
}
    \label{tab:datasets_detailed}
\end{table*}

We noticed some discrepancies between the number of samples reported in the papers and the number of samples in the corresponding raw datasets. Despite following any filtration steps described in the papers, some of the datasets had more samples than the ones in the publicly released version, as indicated in Table \ref{tab:datasets_detailed}. Additionally, the \textit{ArSarcasm-v1}, \textit{Mawqif (Stance Task)}, \textit{Mawqif (Sentiment/Sarcasm Tasks)}, and \textit{ASAD} had 516, 170, 151, 191 samples with less than 3 annotations respectively, that we decided to discard from our analysis.

Conversely, we decided to discard the MLMA dataset \cite{ousidhoum-etal-2019-multilingual} for which the authors shared with us some of the raw annotations files. The number of samples in these files was too small compared to the number of samples in the public dataset with majority-vote labels. We also discarded another dataset, for which there was a significant discrepancy between the released dataset and the paper's description of the dataset.

\section{Code-mixing between MSA and DA}
\label{sec:msa_da_code_mixing}
Researchers distinguish between Modern Standard Arabic (MSA), and Dialectal Arabic (DA) \cite{79835034091e4ec8a327712e27b1b785}. However, MSA and DA do not exist in isolation, and Arabic speakers sometimes code-mix between terms that can be considered to belong to MSA and others considered to be part of a variety of DA. Notably, some terms can be considered to belong to both MSA and a variety of DA, and even using the surrounding context may not be enough for disambiguation \cite{molina-etal-2016-overview}. 

\citet{badawi1973mustawayat} recognizes five levels of Arabic used in Egypt, that can be categorized according to the amount of code-mixing in addition to the dialectness of the terms/phrases used. The \textit{Sentence-ALDi} model, developed by \citet{keleg2023aldi}, estimates the level of dialectness of Arabic sentences, which provides an automatic proxy to distinguish between Arabic sentences according to how they diverge from MSA. We used the \textit{Sentence-ALDi} model to study the relation between the ALDi score and the agreement between the annotators for \ndatasets Arabic datasets.


\section{Discussion about the Analysis}
\label{sec:other_analysis}

As described in \S\ref{sec:methodology}, each dataset's samples were split into 10 bins of equal width according to their respective ALDi scores. Afterward, the correlation between each bin's midpoint ALDi score and the percentage of samples having full agreement \textit{\% full agree} was computed. For each bin, \textit{\% full agree} represents the Maximum Likelihood Estimation (MLE) for the probability that all the annotators agree on the same label for the samples of this bin.

\paragraph{Inability to use Interannotator Agreement metrics for some datasets} Automated metrics such as Fleiss' Kappa \cite{fleiss1971measuring} attempt to measure the Interannotator Agreement (IAA) while accounting for the random agreement/disagreement between annotators. In principle, it might be possible to perform a version of our analysis using Fleiss' Kappa rather than {\it \% full agree} as the dependent variable. However, computing Fleiss' Kappa would require knowledge of the individual annotations for each sample. Such annotations are not available for the ArSAS (Sentiment/Speech Act), DART, and DCD datasets as described in Table
\ref{tab:datasets_detailed}. Since we wanted to include as many datasets as possible, we used {\em \% full agree} instead.

\paragraph{Logistic regression as an alternative analysis tool} Binning the data leads to a loss of analytical information which might impact the results of the analysis, especially if implausible bins' boundaries are used \cite{doi:10.1080/09332480.2006.10722771}. 

Logistic regression with binary outcomes is an alternative analysis that alleviates the limitations of binning. Each sample has a continuous ALDi score as the independent variable, and a binary outcome \textit{Full Annotator Agreement (Yes/No)}. After fitting a logistic regression model to predict the binary outcome, the coefficient of the ALDi variable measures the impact of ALDi on the odds of full agreement. If this coefficient is negative, then the odds of full annotator agreement decrease as the ALDi score increases.

\begin{figure*}[t!]
    \centering
    \scriptsize{\textbf{Sentiment Analysis}}\\
    \begin{flushleft}
        \vspace{-0.65cm}
        \rule{16.5cm}{0.1mm}
        \vspace{-0.48cm}
    \end{flushleft}
    
    \begin{subfigure}[t]{0.25\textwidth}
        \centering
        \includegraphics[trim=0 0.25cm 0 0.2cm, clip]{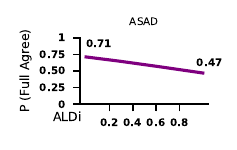}
        \includegraphics[]{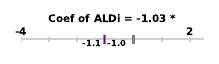}
    \end{subfigure}%
    ~
    \begin{subfigure}[t]{0.25\textwidth}
        \centering
        \includegraphics[trim=0 0.25cm 0 0.2cm, clip]{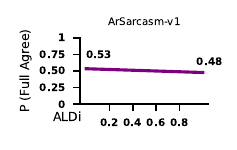}
        \includegraphics[]{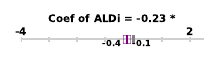}
    \end{subfigure}%
    ~
    \begin{subfigure}[t]{0.25\textwidth}
        \centering
        \includegraphics[trim=0 0.25cm 0 0.2cm, clip]{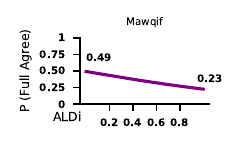}
        \includegraphics[]{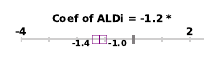}
    \end{subfigure}%
    ~
    \begin{subfigure}[t]{0.25\textwidth}
        \centering
        \includegraphics[trim=0 0.25cm 0 0.2cm, clip]{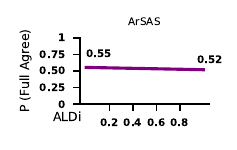}
        \includegraphics[]{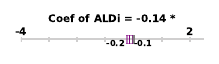}
    \end{subfigure}%

    \vfill
    
    \scriptsize{\textbf{Sarcasm Detection}}\\
    \begin{flushleft}
        \vspace{-0.4cm}
        \rule{12cm}{0.1mm}
    \end{flushleft}

    \begin{subfigure}[t]{0.25\textwidth}
        \centering
        \includegraphics[trim=0 0.25cm 0 0.2cm, clip]{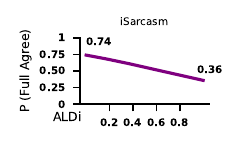}
        \includegraphics[]{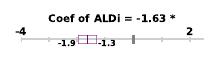}
    \end{subfigure}%
    ~
    \begin{subfigure}[t]{0.25\textwidth}
        \centering
        \includegraphics[trim=0 0.25cm 0 0.2cm, clip]{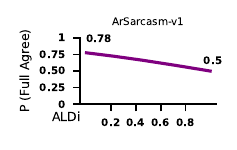}
        \includegraphics[]{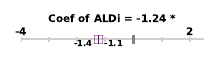}
    \end{subfigure}%
    ~
    \begin{subfigure}[t]{0.25\textwidth}
        \centering
        \includegraphics[trim=0 0.25cm 0 0.2cm, clip]{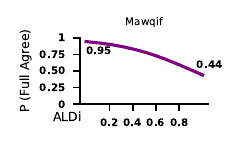}
        \includegraphics[]{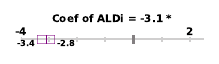}
    \end{subfigure}%
    \begin{subfigure}[t]{0.25\textwidth}
        \centering
        \vspace{-2.4cm}
        \scriptsize{\underline{\textbf{Speech Act Detection}}}
        \includegraphics[trim=0 0.25cm 0 0.2cm, clip]{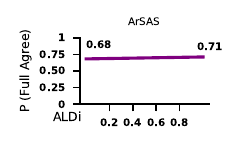}
        \includegraphics[]{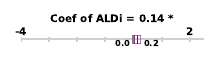}
    \end{subfigure}%
    \vfill

    \scriptsize{\textbf{Offensive Text Classification and Hate Speech Detection}}
    \begin{flushleft}
        \vspace{-0.3cm}
        \rule{12cm}{0.1mm}
    \end{flushleft}

    \begin{subfigure}[t]{0.25\textwidth}
        \centering
        \includegraphics[trim=0 0.25cm 0 0.2cm, clip]{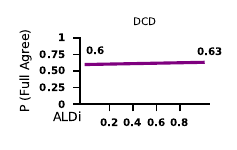}
        \includegraphics[]{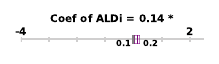}
    \end{subfigure}%
    ~
    \begin{subfigure}[t]{0.25\textwidth}
        \centering
        \includegraphics[trim=0 0.25cm 0 0.2cm, clip]{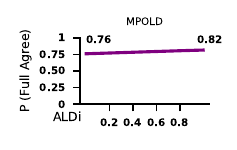}
        \includegraphics[]{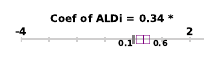}
    \end{subfigure}%
    ~
    \begin{subfigure}[t]{0.25\textwidth}
        \centering
        \includegraphics[trim=0 0.25cm 0 0.2cm, clip]{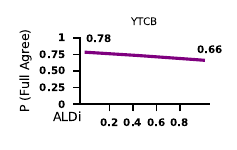}
        \includegraphics[]{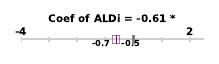}
    \end{subfigure}%
    ~
    \begin{subfigure}[t]{0.25\textwidth}
        \centering
        \vspace{-2.4cm}
        \scriptsize{\underline{\textbf{Stance Detection}}}
        \includegraphics[trim=0 0.25cm 0 0.2cm, clip]{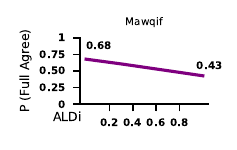}
        \includegraphics[]{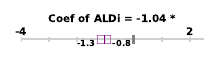}
    \end{subfigure}%
    \vfill
    
    \scriptsize{\textbf{Arabic Dialect Identification}}
    \begin{flushleft}
        \vspace{-0.3cm}
        \rule{12cm}{0.1mm}
    \end{flushleft}

    \begin{subfigure}[t]{0.25\textwidth}
        \centering
        \includegraphics[trim=0 0.25cm 0 0.2cm, clip]{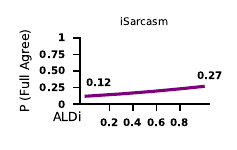}
        \includegraphics[]{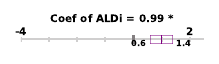}
    \end{subfigure}%
    ~
    \begin{subfigure}[t]{0.25\textwidth}
        \centering
        \includegraphics[trim=0 0.25cm 0 0.2cm, clip]{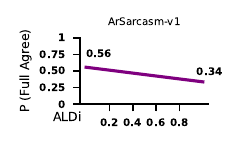}
        \includegraphics[]{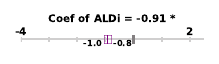}
    \end{subfigure}%
    ~
    \begin{subfigure}[t]{0.25\textwidth}
        \centering
        \includegraphics[trim=0 0.25cm 0 0.2cm, clip]{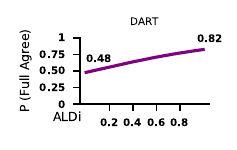}
        \includegraphics[]{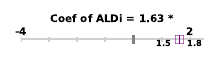}
    \end{subfigure}%
    ~
    \begin{subfigure}[t]{0.25\textwidth}
        \centering
        \phantom{\includegraphics[trim=0 0.25cm 0 0.2cm, clip]{log_reg_plots/DART_dialect.pdf}}
    \end{subfigure}

    \caption{For each dataset, plots show the estimated probability of \textit{full agreement} according to each dataset's fitted logistic regression model. Under each plot, the coefficient of ALDi with its 95\% confidence interval is visualized. Nearly all datasets (marked with *) have confidence intervals that do not include zero, meaning the effect of ALDi is statistically significant at $p< 0.05$. Negative coefficients indicate that higher ALDi scores predict lower agreement.}

    \label{fig:prob_log_reg}
\end{figure*}

Figure~\ref{fig:prob_log_reg} demonstrates the probability of full agreement of each dataset, in addition to the coefficient of the ALDi score with its 95\% confidence interval. For the 8 non-DI datasets with $Coef_{ALDi} < -0.2$, the coefficients can be considered to be statistically significant since the confidence interval does not include zero.

Both analysis tools (correlation analysis and logistic regression) achieve similar results. The same 8 non-DI datasets---ASAD, ArSarcasm-v1 (Sentiment/Sarcasm), Mawqif (Sentiment/Sarcasm/Stance), iSarcasm, and YTCB---have significantly strong negative correlation coefficients as in Figure~\ref{fig:agreement_percentages_new}, and statistically significant coefficients for the ALDi variable which are less than -0.2.
However, binning the data allows for visualizing the \textit{\% full agreement} as a scatter plot, which can reveal whether the relation between ALDi and the agreement is linear or not, in addition to having a visual way for determining how well the best-fitting line models the data.

\paragraph{Impact of data skewness} MSA samples are over-represented in some of the considered datasets. However, this is generally unproblematic for the analysis, so we opted not to discard the MSA samples. For the method described in Section~\ref{sec:methodology}, the samples of each bin are independently used to estimate the MLE of full agreement between annotators. Therefore, the over-representation of MSA samples in some datasets does not impact our analysis.

\section{Trends by Class Label}
\label{sec:detailed_trends}

\begin{figure*}[t]
    \centering
    \begin{subfigure}[t]{0.25\textwidth}
        \centering
        \includegraphics[]{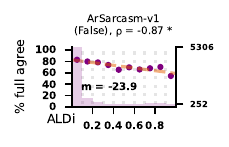}
    \end{subfigure}%
    ~
    \begin{subfigure}[t]{0.25\textwidth}
        \centering
        \includegraphics[]{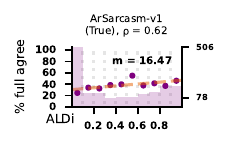}
    \end{subfigure}%

    \vfill

    \begin{subfigure}[t]{0.25\textwidth}
        \centering
        \includegraphics[]{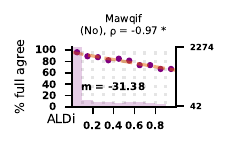}
    \end{subfigure}%
    ~
    \begin{subfigure}[t]{0.25\textwidth}
        \centering
        \includegraphics[]{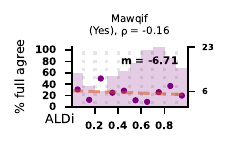}
    \end{subfigure}%

    \vfill

    \begin{subfigure}[t]{0.25\textwidth}
        \centering
        \includegraphics[]{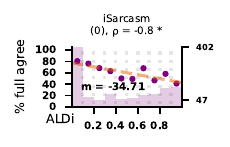}
    \end{subfigure}%
    ~
    \begin{subfigure}[t]{0.25\textwidth}
        \centering
        \includegraphics[]{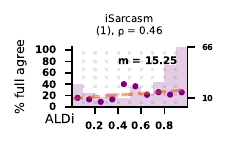}
    \end{subfigure}%
    \caption{The trends for the classes of the Saracasm Detection datasets. Statistically significant correlation coefficients ($\rho$) are marked with *.}
    \label{fig:sarcasm_datasets}
\end{figure*}

\begin{figure*}[t]
    \centering
    \begin{subfigure}[t]{0.25\textwidth}
        \centering
        \phantom{\includegraphics[]{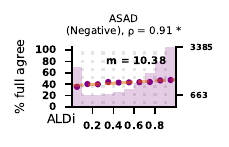}}
    \end{subfigure}%
    ~
    \begin{subfigure}[t]{0.25\textwidth}
        \centering
        \includegraphics[]{new_plots/ASAD_10_sentiment_merged_Negative.pdf}
    \end{subfigure}%
    ~
    \begin{subfigure}[t]{0.25\textwidth}
        \centering
        \includegraphics[]{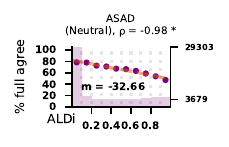}
    \end{subfigure}%
    ~
    \begin{subfigure}[t]{0.25\textwidth}
        \centering
        \includegraphics[]{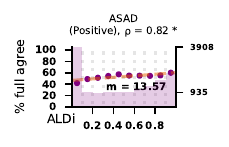}
    \end{subfigure}%

    \vfill

    \begin{subfigure}[t]{0.25\textwidth}
        \centering
        \includegraphics[]{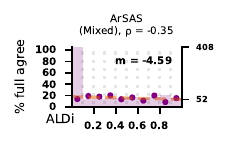}
    \end{subfigure}%
    ~
    \begin{subfigure}[t]{0.25\textwidth}
        \centering
        \includegraphics[]{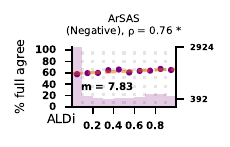}
    \end{subfigure}%
    ~
    \begin{subfigure}[t]{0.25\textwidth}
        \centering
        \includegraphics[]{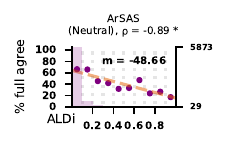}
    \end{subfigure}%
    ~
    \begin{subfigure}[t]{0.25\textwidth}
        \centering
        \includegraphics[]{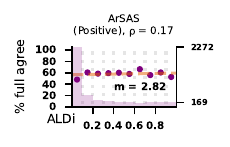}
    \end{subfigure}%
    \vfill
    \begin{subfigure}[t]{0.25\textwidth}
        \centering
        \phantom{\includegraphics[]{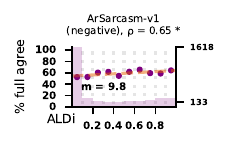}}
    \end{subfigure}%
    ~
    \begin{subfigure}[t]{0.25\textwidth}
        \centering
        \includegraphics[]{new_plots/ArSarcasm-v1_10_sentiment_merged_negative.pdf}
    \end{subfigure}%
    ~
    \begin{subfigure}[t]{0.25\textwidth}
        \centering
        \includegraphics[]{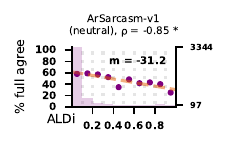}
    \end{subfigure}%
    ~
    \begin{subfigure}[t]{0.25\textwidth}
        \centering
        \includegraphics[]{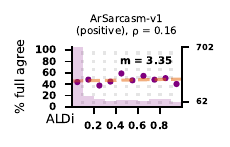}
    \end{subfigure}%
    \vfill
    \begin{subfigure}[t]{0.25\textwidth}
        \centering
        \phantom{\includegraphics[]{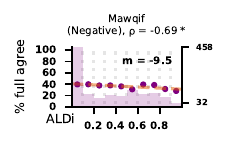}}
    \end{subfigure}%
    ~
    \begin{subfigure}[t]{0.25\textwidth}
        \centering
        \includegraphics[]{new_plots/Mawqif_sarcasm_10_sentiment_merged_Negative.pdf}
    \end{subfigure}%
    ~
    \begin{subfigure}[t]{0.25\textwidth}
        \centering
        \includegraphics[]{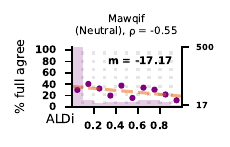}
    \end{subfigure}%
    ~
    \begin{subfigure}[t]{0.25\textwidth}
        \centering
        \includegraphics[]{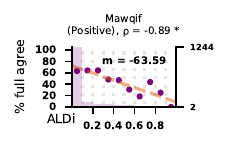}
    \end{subfigure}%
    \caption{The trends for the classes of the Sentiment Analysis datasets. Statistically significant correlation coefficients ($\rho$) are marked with *.}
    \label{fig:sentiment_datasets}
\end{figure*}

\begin{figure*}[t]
    \centering
    \begin{subfigure}[t]{0.25\textwidth}
        \centering
        \includegraphics[]{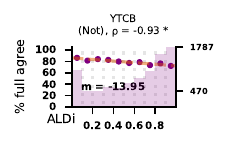}
    \end{subfigure}%
    ~
    \begin{subfigure}[t]{0.25\textwidth}
        \centering
        \phantom{\includegraphics[]{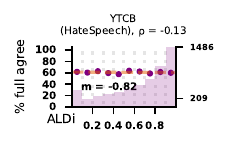}}
    \end{subfigure}%
    ~
    \begin{subfigure}[t]{0.25\textwidth}
        \centering
        \includegraphics[]{new_plots/YouTube_cyberbullying_10_hate_speech_merged_HateSpeech.pdf}
    \end{subfigure}%
    \vfill
    \begin{subfigure}[t]{0.25\textwidth}
        \centering
        \includegraphics[]{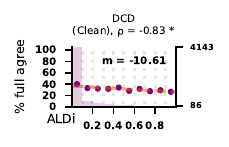}
    \end{subfigure}%
    ~
    \begin{subfigure}[t]{0.25\textwidth}
        \centering
        \includegraphics[]{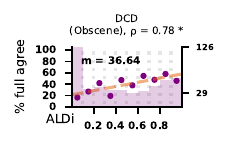}
    \end{subfigure}%
    ~
    \begin{subfigure}[t]{0.25\textwidth}
        \centering
        \includegraphics[]{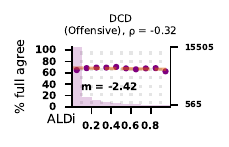}
    \end{subfigure}%
    \vfill
    \begin{subfigure}[t]{0.25\textwidth}
        \centering
        \includegraphics[]{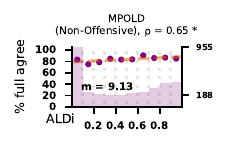}
    \end{subfigure}%
    ~
    \begin{subfigure}[t]{0.25\textwidth}
        \centering
        \phantom{\includegraphics[]{new_plots/MPOLD_10_offensive_merged_Non-Offensive.pdf}}
    \end{subfigure}%
    ~
    \begin{subfigure}[t]{0.25\textwidth}
        \centering
        \includegraphics[]{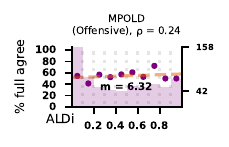}
    \end{subfigure}%
    \caption{The trends for the classes of the Offensive Text Classification and Hate Speech datasets. Statistically significant correlation coefficients ($\rho$) are marked with *.}
    \label{fig:offensive_datasets}
\end{figure*}

\begin{figure*}[hpt]
    \centering
    \begin{subfigure}[t]{0.25\textwidth}
        \centering
        \includegraphics[]{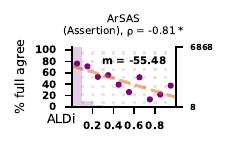}
    \end{subfigure}%
    ~
    \begin{subfigure}[t]{0.25\textwidth}
        \centering
        \includegraphics[]{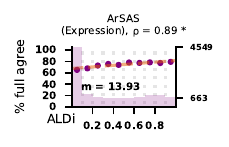}
    \end{subfigure}%
    \caption{The trends for the \textit{Assertion} and \textit{Expression} labels of the ArSAS dataset, which represent 95\% of the dataset samples. Statistically significant correlation coefficients ($\rho$) are marked with *.}
    \label{fig:ArSAS_speech_act_detailed}
\end{figure*}

\begin{figure*}[!t]
    \centering
    \begin{subfigure}[t]{0.25\textwidth}
        \centering
        \includegraphics[]{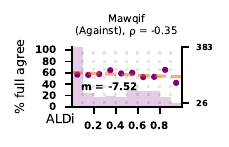}
    \end{subfigure}%
    ~
    \begin{subfigure}[t]{0.25\textwidth}
        \centering
        \includegraphics[]{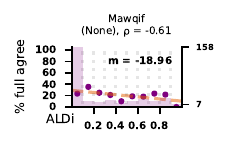}
    \end{subfigure}%
    ~
    \begin{subfigure}[t]{0.25\textwidth}
        \centering
        \includegraphics[]{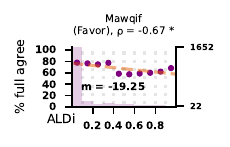}
    \end{subfigure}%
    \caption{The trends for the classes of Mawqif's Stance dataset. Statistically significant correlation coefficients ($\rho$) are marked with *.}
    \label{fig:stance_datasets}
\end{figure*}
As mentioned in \S\ref{sec:detailed_analysis}, Figures~\ref{fig:sarcasm_datasets},~\ref{fig:sentiment_datasets},~\ref{fig:offensive_datasets},~\ref{fig:ArSAS_speech_act_detailed}, and \ref{fig:stance_datasets} visualize the impact of ALDi on the annotator agreement after splitting the samples according to their majority-vote labels. We acknowledge that the number of samples in the bins for some classes is not enough to draw concrete conclusions (e.g., samples with high ALDi scores for the \textit{Neutral} class of the \textit{ArSAS}, and \textit{Mawqif} datasets as per Figure~\ref{fig:sentiment_datasets}).

\section{The Rising Trend of ArSAS}
\label{sec:arsas_trends}
The \textit{ArSAS} dataset stands out as a dataset with a rising trend for the \textit{Speech Act Detection} task and a falling trend for the \textit{Sentiment Analysis} task. Samples of \textit{ArSAS} were jointly annotated for their sentiment and speech act. Despite having 6 different speech acts, which would arguably make speech act detection harder than sentiment analysis, the \textit{Assertion} and \textit{Expression} classes represent 95\% of the samples. Looking at their respective trends shown in Figure~\ref{fig:ArSAS_speech_act_detailed}, the two acts show two different behaviors. Most of the assertive samples have ALDi scores <0.2 (arguably, all are MSA ones). Moreover, the number of \textit{Assestion} samples with high ALDi scores is not enough to estimate the \textit{\% full agree} for their respective bins. Conversely, the \textit{Expression} act shows higher agreement as the ALDi score increases.

The creators of ArSAS noticed that most of the \textit{Assertion} samples were annotated as \textit{Neutral}, while most of the \textit{Expression} samples had polarized sentiment (mostly \textit{Negative}). The annotators might have treated the \textit{Assertion} class as the act for \textit{Objective} sentences, while treating \textit{Expression} as the act for \textit{Subjective} sentences. This is arguably easier than sentiment analysis which might explain why annotators agree more on the Speech Act label than the Sentiment label for the \textit{ArSAS} dataset. Further analysis is required to explain the trends of this dataset.

\end{document}